\documentclass[letterpaper, 10 pt, conference]{ieeeconf}  

\IEEEoverridecommandlockouts                              

\usepackage{hyperref}
\usepackage{graphicx}
\usepackage{amsmath}
\usepackage{cite}
\usepackage{cuted}
\usepackage{caption}
\usepackage{algorithm}
\usepackage{algpseudocode}
\usepackage{multirow}
\usepackage{gensymb}
\usepackage{booktabs}

\IEEEoverridecommandlockouts

\makeatletter
\def\ps@headings{}
\def\@maketitle{%
  \newpage
  \null
  \vskip -2.0em
  \begin{center}%
    {\LARGE \bfseries \@title \par}%
    \vskip 0.5em
    {\normalsize \@author \par}%
    \vskip -1.5em
  \end{center}%
  \par \vskip -2.0em 
}
\makeatother

\title{\LARGE \bf{Planning and Reasoning with 3D Deformable Objects for Hierarchical Text-to-3D Robotic Shaping}}

\author{Alison Bartsch$^{1}$, and Amir Barati Farimani$^{1}$
\thanks{$^{1}$With the Department of Mechanical Engineering,
        Carnegie Mellon University \tt\small \{abartsch, afariman\} @andrew.cmu.edu}}

\begin{document}

\maketitle
\thispagestyle{empty}
\pagestyle{empty}



\begin{strip}
    \centering
    \includegraphics[width=0.99\textwidth]{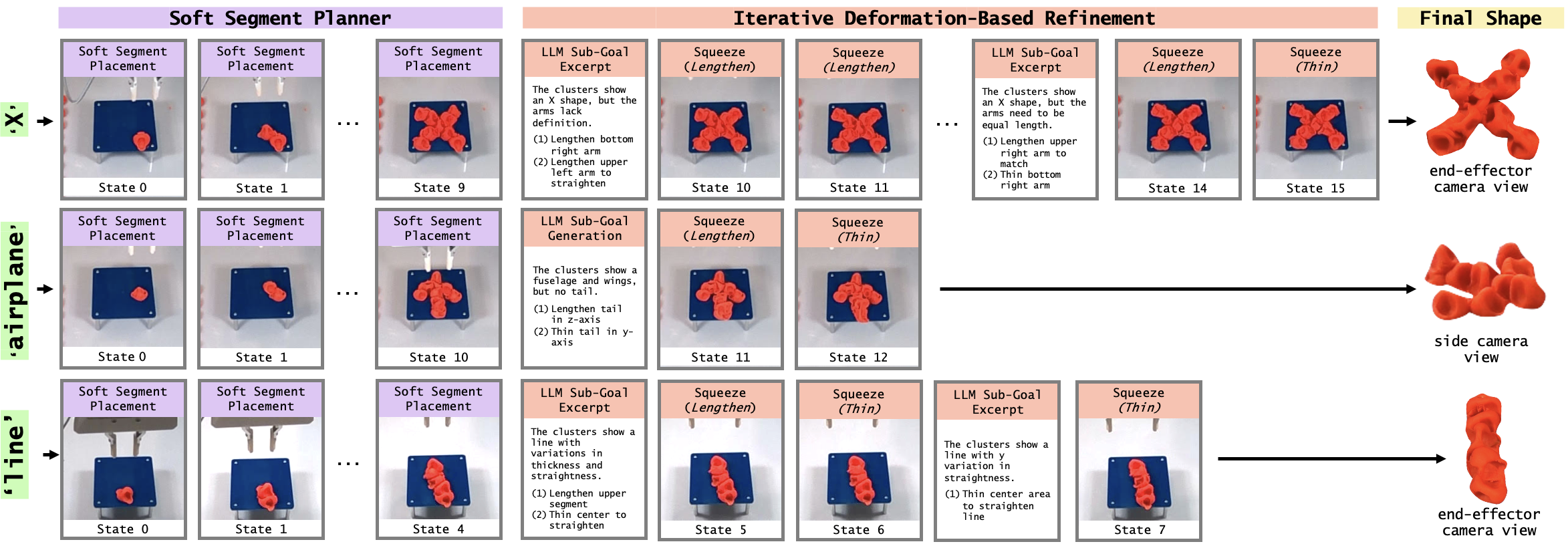} 
    \captionof{figure}{A visualization of the sculpting sequence for our proposed text-to-3D shaping method. Our pipeline first creates a coarse shape in the scene with discrete chunks of clay, and then iteratively refines the shape with deformation-based actions.}
    \label{fig:overview}
\end{strip}


\begin{abstract}

Deformable object manipulation remains a key challenge in developing autonomous robotic systems that can be successfully deployed in real-world scenarios. In this work, we explore the the task of sculpting clay into 3D shapes. We propose the first coarse-to-fine autonomous sculpting system in which the sculpting agent first creates a coarse shape, and then iteratively refines the shape with sequences of deformation actions. We leverage large language models for sub-goal generation, and train a point cloud region-based action model to predict robot actions from the sub-goals. Additionally, our method is the first autonomous sculpting system that is a real-world text-to-3D shaping pipeline without any explicit 3D goals or sub-goals provided to the system. We demonstrate our method is able to successfully create a set of shapes solely from text-based prompting. For experimental videos, human evaluation details, and full prompts, please see our project website: \href{https://sites.google.com/andrew.cmu.edu/hierarchicalsculpting}{https://sites.google.com/andrew.cmu.edu/hierarchicalsculpting}.

\end{abstract}

\section{Introduction}

    As we continue to develop autonomous robotic systems for open-world applications such as surgery \cite{scheikl2022sim, cover1993interactively, wang2022neural}, cooking \cite{shi2023robocook, car2024plato, dikshit2023robochop, liu2022robot, petit2017tracking}, and assistive robotics \cite{sundaresan2022learning, zhang2022learning, jimenez2020perception}, a key and necessary advancement is deformable object manipulation. Deformable objects are prevalent throughout the world, and it is imperative that we build systems that are better able to visualize, reason about and predict behavior of these materials. 3D deformable objects are particularly difficult due to challenges of state representation and the high degrees of freedom for shape prediction and dynamics modeling \cite{shi2024robocraft, bartsch2024sculptbot}. In this work, we investigate the challenges of 3D deformable object manipulation through the lens of the 3D shape creation task.



    Existing autonomous clay shaping methods generally make the key assumption that there is a fixed volume of clay in the workspace, and an explicit 3D goal is provided to guide the planning of a sequence of deformation-based actions to achieve this goal shape \cite{shi2024robocraft, shi2023robocook, bartsch2024sculptbot, bauer2024doughnet, bartsch2024sculptdiff, bartsch2024llm}. In contrast, human sculptors often first roughly add clay to different regions of the sculpture, and slowly refine the shape with many more delicate actions. In this work, we take inspiration from this strategy and propose a coarse-to-fine robotic sculpting approach in which the robot plans a coarse shape and adds discrete chunks of clay to the scene before iteratively refining the shape with small deformation actions. Furthermore, in this work, we abandon the use of explicit 3D point cloud goals, and instead propose a true text-to-3D real-world shaping system. The key contributions of this work are as follows.

    \begin{figure*}
      \centering
      \includegraphics[width=0.875\linewidth]{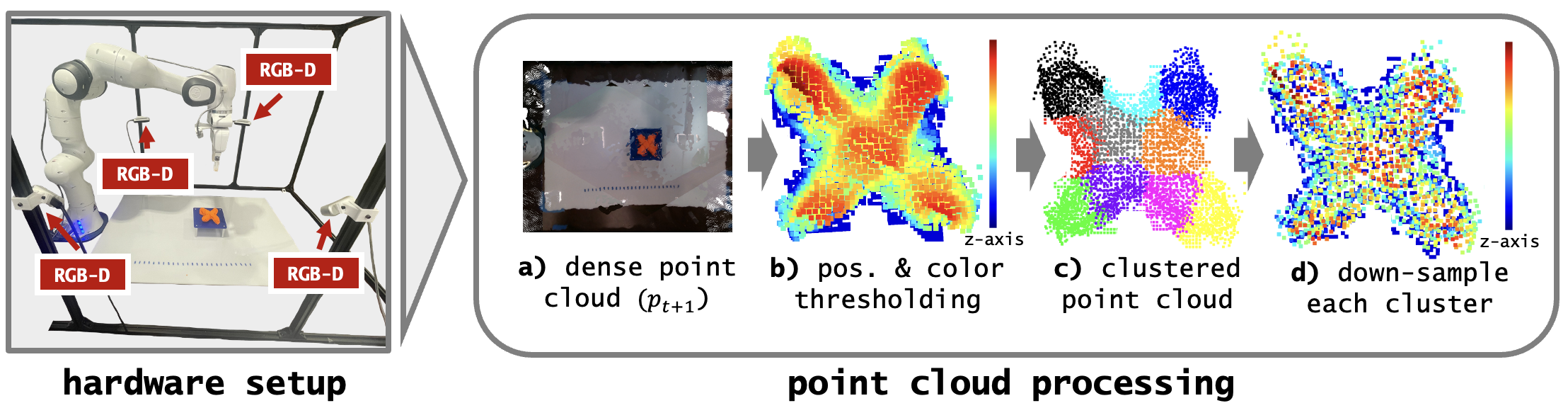}
      \caption{\label{fig:perception} The point cloud processing pipeline first captures a dense point cloud of the robot's workspace (a), then isolates the clay point cloud with position and color thresholding(b), next the point cloud is clustered into 10 regional geometrical patches (c), and finally uniformly down-sampled to ensure each cluster contains an equal number of points.}
    \end{figure*}


    \begin{itemize}
        \item To the best of our knowledge we present both the first autonomous text-to-3D sculpture robotic system without any explicit 3D goals, and the first method combining additive and deformation-based sculpting into a single autonomous pipeline.
        \item We propose a point cloud region-based learned model to predict corresponding deformation actions given a synthetically generated sub-goal.
        \item We leverage Large Language Models (LLMs) for high-level planning of the sculpture shape, as well as sub-goal generation which allows the system to be highly adaptable and exhibit coherent explanations for refinements.
        \item We compare human evaluations to existing shape similarity metrics to evaluate the 3D sculpting task without an explicit 3D goal.
    \end{itemize}




    \section{Related works}

    \textbf{3D Deformable Manipulation: } In recent years, there has been much development in the domain of 3D deformable object manipulation. Dynamics-based 3D deformable object shaping frameworks have been successful for parallel gripper-based clay shaping \cite{shi2024robocraft, shi2023robocook, bartsch2024sculptbot}, with a recent work incorporating topological predictions into the dynamics pipeline \cite{bauer2024doughnet} for more complex dynamics predictions. Recently, researchers presented a recurrent state space model approach to predicting complex dynamics of deformable objects, though the framework struggles for the real-world sculpting task \cite{li2024deformnet}. Alternatively, imitation learning frameworks have been developed for deformation-based sculpting \cite{bartsch2024sculptdiff}, and wheel-based pottery \cite{yoo2024ropotter}. Human demonstrations have also been leveraged for learning inverse dynamics models for the clay rolling task \cite{duan2024human, duan2024plasti}, or for learning to abstract simple dexterous dough manipulation skills \cite{li2023dexdeform}. While these frameworks are effective, human demonstrations can be difficult to collect for these complex tasks and can limit the generalizability of the methods. To alleviate this issue of human demonstrations, researchers leverage a task-agnostic play dataset to learn simple dough manipulation skills with a dexterous hand \cite{yamada2024d}. Large language models (LLMs) have shown promise for task decomposition for the simple clay rolling task \cite{you2024make}, as well as for direct action prediction to create simple shapes in clay \cite{bartsch2024llm}. In this work, we aim to take a different approach to the clay shaping task as compared to the existing literature, combining additive and deformation-based sculpting techniques together to first create a general 3D shape, and then iteratively refine the sculpture with small grasp actions.

    \textbf{LLMs as Robotic Planners: } LLMs and Vision Language Models (VLMs) have been demonstrated to contain useful work knowledge for directly generating successful plans for a variety of simple robotics tasks, from cleaning a household environment \cite{wu2023tidybot}, to tabletop manipulation tasks \cite{mai2024vlm}, and even creative tool use \cite{xu2023creative}. Following these results, researchers found that with deliberate state and action space considerations, LLMs and VLMs are able to generate quality manipulation plans for deformable manipulation tasks including cloth folding \cite{raval2024gpt}, and 2D dough shaping \cite{bartsch2024llm}. More sophisticated methods to better ground the generated LLM plans can result in better and more reliable outputs. In PIVOT, researchers present an iterative visual prompting strategy which frames action generation as a sequence of visual question-answering for VLMs to better generate action plans for a variety of robotic tasks \cite{nasiriany2024pivot}. Incorporating human feedback into the system can also improve performance and ground its output \cite{liang2024learning, wang2024grounding}. Alternatively, incorporating additional learned models can improve performance of these LLM-based approaches, such as adding a learned grasp affordance model to ensure LLM-generated plans for tool use are able to consider grasp affordances \cite{car2024plato}. In this work we aim to leverage the text-based prompting and useful world knowledge of LLMs to design a generalizable and adaptable text-to-3D clay shaping system.

\section{Methodology}
    There are four main components of the proposed hierarchical text-based sculpting system: the perception module (section \ref{sec:perception}), the soft segment planner (section \ref{sec:planner}), the sub-goal generation module (section \ref{sec:subgoal}), and the learned action model (section \ref{sec:model}).

    \subsection{Perception}
    \label{sec:perception}

    While our pipeline does not require an explicit 3D goal shape, we still need to accurately capture the 3D shape of the clay as it is being sculpted. A visualization of the perception pipeline is shown in Figure \ref{fig:perception}. Our hardware setup contains four Intel RealSense D415 cameras mounted to provide a complete 3D reconstruction of the robot's workspace. A point cloud is captured before and after each additive or deformation-based action the robot executes. After the dense point cloud of the scene is captured, the clay point cloud is isolated with position and color-based thresholding. However, this simple thresholding technique could be replaced with alternative segmentation systems, such as SAM \cite{kirillov2023segment} in cases where color-based thresholding would not be possible. After the clay point cloud is isolated, the point cloud is clustered into 10 regional patches with k-nearest-neighbors \cite{cover1967knn}. We chose to represent the 3D geometry of the clay at a local cluster level to enable the learned action model (Section \ref{sec:model}) to better generalize to unseen global clay geometries. Finally, the point cloud is uniformly downsampled such that each cluster contains 256 points.

    \subsection{Soft Segment Planner}
    \label{sec:planner}


    We propose a coarse-to-fine shape generation approach in which the robot first adds segments of clay to the scene, and then iteratively refines the shape with a sequence of deformation-based actions. The soft segment planner is the system that first places the discrete segments of clay in the scene given a text-based goal shape. We leverage the LLM Gemini 1.5 \cite{team2024gemini} and a text-to-point-cloud model, Point-E \cite{nichol2022point}, to generate the segment placement plan, with the state represented as a discrete 5x5x5 3D grid, where each cell has a dimension of 1.5cm. Given the text-based goal prompt, the LLM determines if it would like to query Point-E for help with an initial guess of the occupancy grid (initialized as $s_0$), or start with a blank grid. We found that this improves the planner's performance for more complex 3D shapes, as the LLM alone can hallucinate. Based on the initialized $s_0$ the LLM iteratively adds or removes cells in the 3D grid until the goal shape has been reached, creating an ordered set of soft segments to place in the scene. The LLM must consider the final placement constraints, including reachability and gravity. The details of the planner are described in Algorithm \ref{alg:planner}. The soft segment planner iteratively generates the plan offline, then executes the clay placement sequence by picking equally sized clay balls from a pre-defined location and placing them in the scene. For full details of the prompts provided to the LLM, please visit our project website.

    \begin{algorithm}
        \caption{Soft Segment Planner}
        \label{alg:planner}
    \begin{algorithmic}
        \State\textbf{Input:} Text-based shape prompt ($p$), Initial scene state ($s_0$), Planner LLM ($\sigma$), Removal LLM ($\phi$), Termination LLM ($\theta$), Assistance LLM ($\gamma$), Point-E Model ($\zeta$)
        \State\textbf{Output: } Ordered set of soft cell segments to place in scene $OS = \{OS_0, OS_1, ... , OS_N\}$
        \State\textbf{Initialize: } $OS = \{\}$
        \If{$\gamma(p) = $ False} \Comment{if LLM needs help with shape}
            \State $\tau_{s_0} = \zeta(p)$ \Comment{get occupancy grid of Point-E cloud}
            \State $s_0' = \sum \tau_{s_0}$ \Comment{update initial grid}
            \State $OS' = \tau_{s_0}$ \Comment{initialize ordered placement}
        \EndIf
        \While{$\theta(s_t) = $ False}
            \If{iter $\%$ 2 = 0} \Comment{if even number}
                \State $\tau_{\sigma} = \sigma(p,s_t)$ \Comment{$\sigma$ predicts cells to add}
                \State $s_{t+1} = s_t + \sum \tau_{\sigma}$ \Comment{Update discrete scene cells}
                \State $OS' = OS \cup \tau_{\sigma}$ \Comment{Add cells to plan}
            \Else{}
                \State $\tau_{\phi} = \phi(p, s_{t+1})$ \Comment{$\phi$ predicts cells to remove}
                \State $s_{t+1} = s_t - \sum \tau_{\phi}$ \Comment{Update discrete scene cells}
                \State $OS' = OS \setminus OS \cap \tau_{\phi}$ \Comment{Remove cells from plan}
            \EndIf
            \State iter = iter + 1
        \EndWhile
        \State \textbf{Return:} $OS$ 
    \end{algorithmic}
    \end{algorithm}

    \subsection{LLM-Based Sub-Goal Generation}
    \label{sec:subgoal}




    After the soft segment planner has finished, there is a coarse shape of clay in the scene with variable volume. To improve the quality of the final shape, the robot must now apply a sequence of refinement actions. However, without an explicit 3D goal to compare with the current clay state, it is very difficult to plan the sequence of refinement actions necessary to improve the shape quality. In this section, we propose an LLM-based sub-goal generator that given the text goal prompt and the current 3D point cloud of the clay, will generate sub-goals leveraging an API of point cloud modification functions we created. Generating intermediate sub-goals for deformable object shape creation is a very difficult problem. Past work has avoided this challenge by assuming that the sub-goals are provided by human demonstrations \cite{shi2023robocook}. We chose to leverage the world knowledge of LLMs to enable sub-goal generation without relying on any human-collected data for each task. The LLM sub-goal generator is provided with a text-based representation of the sparse point cloud, listing out each point in x,y,z for each of the clusters. The LLM must determine which clusters need to be modified, and how to improve the final goal shape. We use Gemini 1.5 \cite{team2024gemini} as our LLM with no further finetuning. Each point cloud modification requires the LLM to select the specific cluster to modify, the direction to apply the modification, and a weighting term between 0 and 1 to indicate how extreme of a modification to apply. After the LLM selects these parameters, we pass them directly into the modification function to transform the selected point cloud. The point cloud modification functions are described below: 


    \begin{itemize}
        \item \textit{'lengthen': } Increase the spacing between points along the direction provided to the function. The space added is proportional to the weighting term. 
        \item \textit{'shorten': } Reduce the spacing between points along the direction provided to the function. The space removed is proportional to the weighting term.
        \item \textit{'flatten': } Reduce the z-values of the points in the point cloud proportional to the weighting term, but with a floor to the lowest z value in the cluster.
        \item \textit{'thin': } Reduce the spacing between points perpendicular to the x,y direction provided to the function. The space removed is proportional to the weighting term.
    \end{itemize}



    We ensure that the point cloud is further downsampled to 50 points per cluster before querying the LLM to ensure it can be represented in text form as 'point at (x,y,z)'. Furthermore, we enforce that the cluster labels are always ordered from the minimum centroid z to the maximum centroid z. For any centroids that are at the same z height, the clusters are then ordered from minimum x and y to maximum x and y. We found that consistency in the ordering of the clusters was key to the success of this module, particularly in reducing hallucinations. This enables the LLM to always know that the first cluster provided will be at the bottom of the shape, and the last cluster provided will be at the top. This reduces the complexity of the task for the LLM, as it does not need to reason about the relative x,y,z positions of each cluster to get this information. For full access to the prompts used, please visit our project website.

    



    \begin{figure*}
      \centering
      \includegraphics[width=0.975\linewidth]{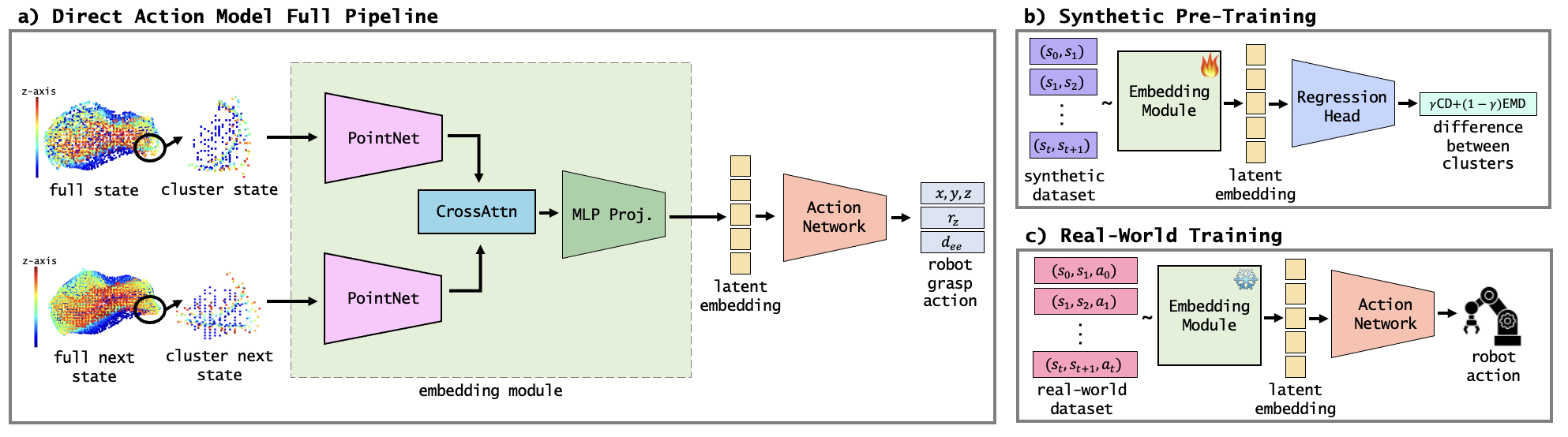}
      \caption{\label{fig:model} a) The full direct action model pipeline with a cluster-based siamsese PointNet embedding network. b) The synthetic pre-training strategy. c) The real-world action finetuning strategy.}
      \label{figurelabel}
    \end{figure*}

    \subsection{Point Cloud Action Model}
    \label{sec:model}

    A key contribution of this work is the development of a direct observation-to-action learned sculpting model that, given a current state and goal state, is able to propose the grasp action to achieve this difference. The action space is parameterized as the x,y,z position of the end-effector, the rotation of the end-effector about the z-axis, and the distance between the fingertips. A direct observation to action model circumvents the need to predict the dynamics of the clay itself. However, real-world data collection for deformation-based sculpting is quite time consuming, and a key challenge of this approach is how to develop a model that can successfully predict actions without requiring a large quantity of real-world data to train. Furthermore, given that we are collecting point cloud observations before and after the entire grasp is complete, there are very large state differences between the observed and goal point clouds making accurate action prediction more difficult. 

    In this work, we propose a region-based action model. The motivation for working at the region level as opposed to a complete point cloud of the entire clay is that it prevents the learned model from overfitting to the specific global geometries of the clay in the training data, allowing the model to much better generalize to unseen global geometries. A visualization of the model details is shown in Figure \ref{fig:model}. Our direct action model employs a siamese architecture, in which there are two branches of PointNet \cite{qi2017pointnet} to embed the current state cluster and sub-goal cluster respectively. We take the output of the final feature transform layer from PointNet and concatenate those regional features ($N \times 64$) with the final max pool global feature ($1 \times 1024$) to reach a latent embedding of $N \times 1088$. This latent embedding contains regional and global feature information for the state and next state clusters respectively. These two latent embeddings are weighted with a cross-attention layer, and then projected into a lower dimensional latent embedding. Finally this latent embedding is used to predict the 5D grasp action with an MLP action network. We train this cluster-based direct action model in two stages, first pre-training the point cloud embedding with a synthetically generated dataset, and then freezing the point cloud embedding models to train the action prediction network on a small real-world dataset of state, next state and action tuples. To generate the synthetic dataset, we reuse the point cloud modification API from Section \ref{sec:subgoal} by iteratively randomly sampling the functions to modify a small set of real-world point clouds. The pre-training synthetic dataset contains 4000 state and next state pairs. The pre-training objective was a weighted prediction of the Chamfer Distance \cite{rubner2000earth} and Earth Mover's Distance \cite{fan2017point} between the the two point clouds ($\gamma CD(s_t, s_{t+1}) + (1 - \gamma) EMD(s_t, s_{t+1})$). We found this pre-training objective outperformed alternative BYOL-style pre-training \cite{grill2020bootstrap}, and we hypothesize this is because predicting the point-wise differences between the two point cloud clusters is very useful for the downstream task of predicting the action to cause the geometrical change between the two states.

\section{Experiments and Results}

    To evaluate our proposed method, we first analyze the action prediction model compared to a wide range of baselines in Section \ref{sec:action_analysis}. We then analyze the effectiveness of quantitative metrics to evaluate the text-to-3D objective in Section \ref{sec:clip_analysis}. We then discuss our human evaluation methodology in Section \ref{sec:human_eval} Finally, we present the results of the full text-to-3D pipeline in Section \ref{sec:texto3d}, and a discussion of semantically tuning the shapes in Section \ref{sec:semantic}.

    \subsection{Point Cloud Cluster-Based Action Model}
    \label{sec:action_analysis}

     \begin{table*}[]
\caption{\textbf{Action model evaluations} on real-world validation and test datasets, and synthetically generated sub-goals. }\label{tab:action_model_performance} 
\centering
\begin{tabular}{@{\extracolsep{\fill}}llllllll}
    \toprule
    & & \multicolumn{2}{l}{\textbf{--------------- Real-World Goals ---------------}}  & \multicolumn{3}{l}{\textbf{-------------------------- Synthetic Goals -------------------------}} \\
    \midrule
    & & Val. MSE ($\times 10^{-2}$)   & Test MSE ($\times 10^{-2}$) & CD ($\times 10^{-3}$) & EMD ($\times 10^{-3}$)  & HD ($\times 10^{-2}$) \\
    \hline
    \midrule
    
    \centering \multirow{5}{*}{\textbf{Flow}}   & VINN & 4.6738 $\pm$ 0.4471  & 6.2036 $\pm$ 0.6041  & 4.9414 $\pm$ 1.2724  & 5.3373 $\pm$ 1.6672  & 2.8986 $\pm$ 0.6746 \\
    & NN-greedy & 5.2746 $\pm$ 0.6096  & 6.6528 $\pm$ 0.5129  & \textbf{4.0909 $\pm$ 0.6481} & \textbf{3.9068 $\pm$ 0.5177} & \textbf{2.5559 $\pm$ 0.5827}  \\
    & DM frozen & \textbf{0.3172 $\pm$ 0.0009} &\textbf{0.4941 $\pm$ 0.0126} & 5.5814 $\pm$ 1.3489 & 5.8294 $\pm$ 1.6437        & 3.0257 $\pm$ 0.7006 \\
    & DM unfrozen & 0.3230 $\pm$ 0.0033  & 0.5283 $\pm$ 0.0193  & 4.9325 $\pm$ 0.9413  & 5.3881 $\pm$ 1.0089  & 2.8461 $\pm$ 0.6242 \\
    & DM end-to-end & 0.3275 $\pm$ 0.0029 & 0.5532 $\pm$ 0.0228  & 4.8012 $\pm$ 1.1980 & 5.0530 $\pm$ 1.7448   & 2.9253 $\pm$ 0.3432  \\
    \midrule

    \multirow{5}{*}{\textbf{Cluster}} & VINN & 1.9841 $\pm$ 0.5037 & 4.0220 $\pm$ 0.6039 & 6.1351 $\pm$ 1.7492 & 9.0288 $\pm$ 3.9392 & 3.5216 $\pm$ 0.7914      \\
    & NN-greedy & 3.4276 $\pm$ 0.8817 & 6.4249 $\pm$ 0.6374 & \textbf{4.3366 $\pm$ 1.0291} & \textbf{4.5303 $\pm$ 0.8991} & \textbf{2.4911 $\pm$ 0.6684} \\
    & DM frozen & \textbf{0.0973 $\pm$ 0.0109} & \textbf{0.1294 $\pm$ 0.0091} & 5.3883 $\pm$ 1.4179 & 5.7729 $\pm$ 2.1100        & 3.0440 $\pm$ 0.5961  \\
    & DM unfrozen & 0.1959 $\pm$ 0.0350 & 0.2297 $\pm$ 0.0657 & 5.0273 $\pm$ 0.8747 & 6.2541 $\pm$ 1.9443     & 2.9237 $\pm$ 0.8124  \\
    & DM end-to-end & 0.1750 $\pm$ 0.0549 & 0.2274 $\pm$ 0.0517 & 4.8009 $\pm$ 1.9682 & 5.6227 $\pm$ 3.2461    & 2.7793 $\pm$ 1.1437 \\
    \midrule

    \multirow{5}{*}{\textbf{Full}} & VINN      & 3.1002 $\pm$ 0.0814 & 4.4353 $\pm$ 0.1939 & 5.6875 $\pm$ 2.4857 & 7.1795 $\pm$ 5.5552 & 2.9680 $\pm$ 0.8701      
    \\
    & NN-greedy & 6.5680 $\pm$ 0.6836 & 8.2210 $\pm$ 0.4459 & \textbf{4.2591 $\pm$ 1.0456} & \textbf{4.1816 $\pm$ 1.0240 }       & \textbf{2.3224 $\pm$ 0.7426}  \\
    & DM frozen  & \textbf{0.3827 $\pm$ 0.0183}  & \textbf{0.4506 $\pm$ 0.0213}  & 4.5189 $\pm$ 1.0245 & 5.0411 $\pm$ 1.4136     & 2.3503 $\pm$ 0.7216 \\
    & DM unfrozen   & 0.6211 $\pm$ 0.1155  & 0.5499 $\pm$ 0.0192 & 5.3221 $\pm$ 1.3302 & 6.1727 $\pm$ 2.4955   & 2.8677 $\pm$ 0.6568  \\
    & DM end-to-end & 0.4344 $\pm$ 0.0116 & 0.5014 $\pm$ 0.0267 & 5.1410 $\pm$ 1.4551 & 5.2690 $\pm$ 1.7275    & 2.9970 $\pm$ 0.9365   \\
    \midrule
    
    ---------- & Random & 5.7768 $\pm$ 0.6608 & 7.7623 $\pm$ 0.4428 & 5.1962 $\pm$ 3.2386 & 6.1937 $\pm$ 4.1658 & 3.4821 $\pm$ 2.7646 \\
    \bottomrule
\end{tabular}
\end{table*}

    To evaluate the quality of the point cloud action model, we withhold a test set of real-world state, action, next state data points. We report the mean squared error (MSE) between the predicted and ground truth actions given the state/next state pairs. We compare our proposed cluster-based point cloud action model to a variety of baselines. Firstly, to explore the benefits and drawbacks of different state representation choices, we compare the proposed cluster-based representation to a complete point cloud representation, and a flow field for the specific cluster. To generate the flow field, we take the cluster-based state/next state pair and employ coherent point drift correspondence algorithm \cite{myronenko2010point} to find matching points. Furthermore, for each state representation, we investigate an array of possible action model frameworks. These variants are listed below.
    \begin{itemize}
        \item \textbf{VINN-style: } We implement Visual Imitation Through Nearest Neighbors (VINN) \cite{pari2021surprising}, with the frozen embedding module pre-trained on the synthetic dataset. 
        \item \textbf{NN-greedy: } Execute nearest neighbor's action in the latent space of the frozen embedding module pre-trained on the synthetic dataset.
        \item \textbf{pre-trained frozen direct model (DM frozen): } The embedding is pre-trained on the synthetic dataset, and then is frozen while the latent action model is trained on the real-world dataset.
        \item \textbf{pre-trained unfrozen direct model (DM unfrozen): } The embedding is pre-trained on the synthetic dataset, and is further trained while the latent action model is trained on the real-world dataset.
        \item \textbf{end-to-end direct model (DM end-to-end): } There is no pre-training, the embedding and latent action model are both trained only on the real-world dataset.
    \end{itemize}


    Beyond direct MSE prediction metrics, we also need to evaluate the performance of our action model on synthetically generated sub-goals, as this is how the action model will be integrated into the overall multi-level sculpting framework. However, we have no ground truth actions for these synthetic sub-goals. Instead, we conduct a series of hardware experiments, and report the Chamfer Distance (CD) \cite{rubner2000earth}, Earth Mover's Distance (EMD)\cite{fan2017point}, and Hausdorff Distance (HD) \cite{taha2015efficient} between the synthetically generated sub-goal complete point cloud and the complete real world point cloud after the proposed grasp was executed. For these experiments, we conduct 10 experimental runs for each model variant and initialize the clay 2 times each for the following shapes that were unseen during training: 'circle', 'triangle', 'column', 'star', and 'cube'. The results of the experiments are shown in Table \ref{tab:action_model_performance}. From our experiments, we find that the cluster-based representation significantly outperforms the flow and complete point cloud representations in terms of MSE. We surmise that this is likely because the test set contains actions for unseen shapes. The full point cloud representation struggles to predict local actions for unseen global geometries. Whereas the cluster-level model only relies on the regional geometry to make action predictions, allowing it to better handle unseen clay shapes. While the flow-based model is also a cluster-level representation, it contains a substantial amount of noise due to the challenges of deformable point correspondence which we believe is hurting that representation's performance. For model type, the pre-trained frozen direct model performs the best with action prediction. We hypothesize that by freezing the point cloud embedding model after training on a much larger synthetic dataset, we are preventing the PointNet model from over-fitting to the substantially smaller real-world dataset when training the latent action prediction model. For the hardware experiments with synthetic goals, the differences between the representations and model types are much less distinct. Unfortunately small differences in the CD/EMD/HD metrics do not fully quantify the differing behavior of the various models. The best performing model on the synthetic goal task as the NN-greedy method. However, this method primarily chose actions that missed the clay entirely or did not interact with the selected cluster of interest for the synthetic sub-goal. Some of the changes proposed for the synthetic sub-goals are quite small variations, in which not interacting with the clay can still yield relatively small CD/EMD/HD results comparable to those of the direct models that are correctly interacting with the modified cluster. Due to the ambiguity in the synthetic goal experiments and the reliability of the MSE metrics to quantify action prediction quality, we have selected the cluster-based pre-trained frozen direct model for the rest of our full pipeline experiments.

    \subsection{Analysis of CLIP Scores for Clay Sculpting}
    \label{sec:clip_analysis}

    \begin{figure}
      \centering
      \includegraphics[width=1.0\linewidth]{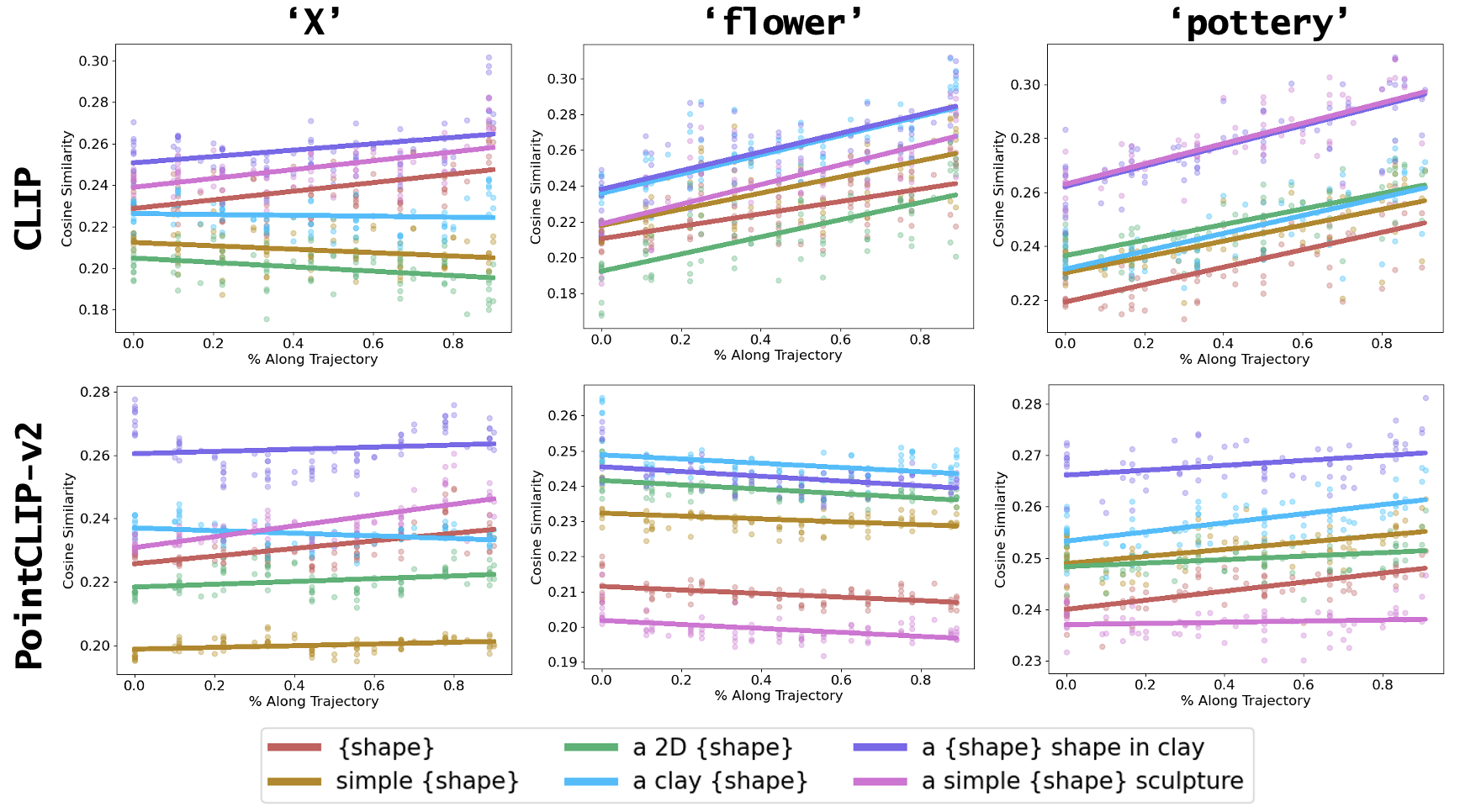}
      \caption{\label{fig:clip} Scatter plot with line of best fit for the CLIP and PointCLIP-v2 cosine similarity of text and image/point cloud embeddings of 10 human trajectories creating each shape in clay. The line of best fit's slope for each shape and prompt shows how well the CLIP or PointCLIP-v2 score correlates with our human oracle-created shapes and varying prompts.}
      \label{figurelabel}
    \end{figure}

    While we were able to use concrete quantitative metrics to evaluate the performance of the action model separately, for the complete pipeline evaluations we cannot use these common shape similarity metrics as we do not have any ground truth 3D goals. In this section, we aim to analyze if existing text and 2D/3D pre-trained models are able to reasonably quantify the results for the clay sculpting task. We specifically investigated the performance of CLIP \cite{radford2021learning} and PointCLIP-v2 \cite{zhu2023pointclip} for a set of human-created shapes in clay: X, flower, and pottery. We plot the CLIP and PointCLIP-v2 embeddings of the states of the clay along the trajectory as a human creates the goal shapes, shown in Figure \ref{fig:clip}. We hypothesize that if these metrics are able to capture text goal and 3D clay sculpture similarity, the cosine similarity scores between the text and image or point cloud embeddings should increase over the human trajectory, as the final state is more similar to the text-based goal than the initial state. For these experiments, we find a high variability in correlation between the CLIP cosine similarity scores of the sculpture and the step in the sculpting trajectory (for the best performing prompt 'a $\{$shape$\}$ shape in clay', $R^{2}_X=0.15$, $R^{2}_{flower} = 0.55$, $R^{2}_{pottery} = 0.60$), while there is a lower correlation with the PointCLIPv2 scores (for the best performing prompt prompt '$\{$shape$\}$', $R^{2}_X = 0.02$, $R^{2}_{flower} = 0.19$, $R^{2}_{pottery} = 0.003$). Furthermore, while the final state in the human trajectories consistently had higher CLIP scores, the magnitude of the change in the score between the initial and final state is very small, making it difficult to make direct comparisons of CLIP scores. These results demonstrate that CLIP scores are not sufficient as the sole quantitative evaluation metric for our text-to-3D pipeline. Therefore, for our text-to-3D experiments, we will supplement the CLIP scores with human evaluations. We hypothesize that CLIP embeddings are not always accurate for this task, as we are creating rough sculptures of a goal object, meaning there is a large difference visually between the real object and the sculpture. Additionally, PointCLIPv2 is pre-trained on a substantially smaller dataset of point clouds, and may not have been trained for these specific shapes. Furthermore, the CLIP embeddings with 'clay' in the prompt had a stronger positive correlation. An interesting avenue of future research would be to develop better quantitative metrics for the task of real-world clay shaping without explicit 3D goal shapes.

    \subsection{Human Survey}
    \label{sec:human_eval}

    Given the variability of the results of the CLIP analysis across shapes and prompts, we need to provide an additional metric to quantify the performance of our proposed system. We conducted a survey of 25 people, who were provided an image of the shape created by the robot and a human oracle (the order in which the shapes were seen was randomized). For each shape, respondents were asked to rate on a scale of 1-5 answering the following  questions: (1) \textit{How well does this shape match the prompt $\{$shape$\}$?}, (2) \textit{What would you rate the quality of this shape?}. With these two questions, we aim to collect evaluation data for how well the shapes match the prompts, as well as the overall quality or skill level to create the shape. For the first question, the scale was 1 corresponding to "not at all", and 5 corresponding to "extremely". For the second question, the scale was 1 corresponding to "poor" and 5 corresponding to "excellent". As an additional metric, we ask an LLM question (1) to rate the final point cloud with respect to the text prompt.

    \begin{table*}[]
\caption{\textbf{Full Pipeline Evaluations}. The Image and text CLIP scores are between 0 and 1. The human evaluations (denoted with H.) and LLM ratings are between 1 and 5.}\label{tab:pipeline_eval} 
\centering
\begin{tabular}{@{\extracolsep{\fill}}llllllllll}
    \toprule
    & & \multicolumn{4}{l}{\textbf{------------------------------ Full Pipeline ------------------------------}}  & \multicolumn{4}{l}{\textbf{----------------------------- Human Oracle -----------------------------}} \\
    \midrule
    & & CLIP $\uparrow$  & LLM $\uparrow$  & H. Prompt $\uparrow$ & H. Quality $\uparrow$ & CLIP $\uparrow$ & LLM $\uparrow$ & H. Prompt $\uparrow$ & H. Quality $\uparrow$ \\
    \hline
    \midrule

    & \textbf{X} & 0.278 $\pm$ 0.014 & 2.33 $\pm$ 0.47 & 4.63 $\pm$ 0.70 & 3.75 $\pm$ 0.97 & 0.295 $\pm$ 0.008 & 2.67 $\pm$ 0.47 & 4.15 $\pm$ 0.53 & 3.69 $\pm$ 0.82 \\
    & \textbf{line} & 0.271 $\pm$ 0.009 & 2.67 $\pm$ 0.47 & 4.38 $\pm$ 0.70 & 3.38 $\pm$ 0.99 & 0.289 $\pm$ 0.002 & 3.33 $\pm$ 0.47 & 4.62 $\pm$ 0.63 & 4.15 $\pm$ 0.77 \\
    & \textbf{flower} & 0.294 $\pm$ 0.004 & 3.00 $\pm$ 0.00 & 3.13 $\pm$ 1.05 & 2.88 $\pm$ 0.78 & 0.302 $\pm$ 0.021 & 3.00 $\pm$ 0.00 & 4.31 $\pm$ 0.61 & 4.00 $\pm$ 1.11 \\
    & \textbf{column} & 0.270 $\pm$ 0.003 & 3.33 $\pm$ 0.47 & 2.88 $\pm$ 1.17 & 2.63 $\pm$ 1.22 & 0.306 $\pm$ 0.005 & 2.33 $\pm$ 0.47 & 3.31 $\pm$ 1.14 & 3.00 $\pm$ 1.04 \\
    & \textbf{pyramid} & 0.275 $\pm$ 0.009 & 3.00 $\pm$ 0.00 & 1.62 $\pm$ 0.74 & 1.54 $\pm$ 0.93 & 0.289 $\pm$ 0.007 & 3.00 $\pm$ 0.00 & 4.75 $\pm$ 0.42 & 3.75 $\pm$ 0.66 \\
    & \textbf{airplane} & 0.276 $\pm$ 0.009 & 3.00 $\pm$ 0.00 & 2.31 $\pm$ 0.82 & 1.85 $\pm$ 0.77 & 0.306 $\pm$ 0.005 & 2.67 $\pm$ 0.47 & 3.88 $\pm$ 0.93 & 3.13 $\pm$ 1.27 \\
    & \textbf{chair} & 0.297 $\pm$ 0.013 & 2.00 $\pm$ 0.00 & 1.85 $\pm$ 0.77 & 1.50 $\pm$ 0.76 & 0.327 $\pm$ 0.007 & 2.67 $\pm$ 0.47 & 3.00 $\pm$ 0.71 & 2.75 $\pm$ 0.97 \\
    & \textbf{pottery} & 0.275 $\pm$ 0.018 & 3.00 $\pm$ 0.00 & 1.39 $\pm$ 0.63 & 1.39 $\pm$ 0.63 & 0.297 $\pm$ 0.002 & 3.00 $\pm$ 0.00 & 4.02 $\pm$ 0.95 & 3.00 $\pm$ 1.23 \\
    \midrule
    & \textbf{all shapes} & 0.279 $\pm$ 0.014 & 2.79 $\pm$ 0.50 & 2.54 $\pm$ 1.37 & 2.18 $\pm$ 1.20 & 0.301 $\pm$ 0.015 & 2.83 $\pm$ 0.47 & 4.02 $\pm$ 0.95 & 3.50 $\pm$ 1.11 \\
    \bottomrule
    
\end{tabular}
\end{table*}

    \begin{figure*}
      \centering
      \includegraphics[width=1.0\linewidth]{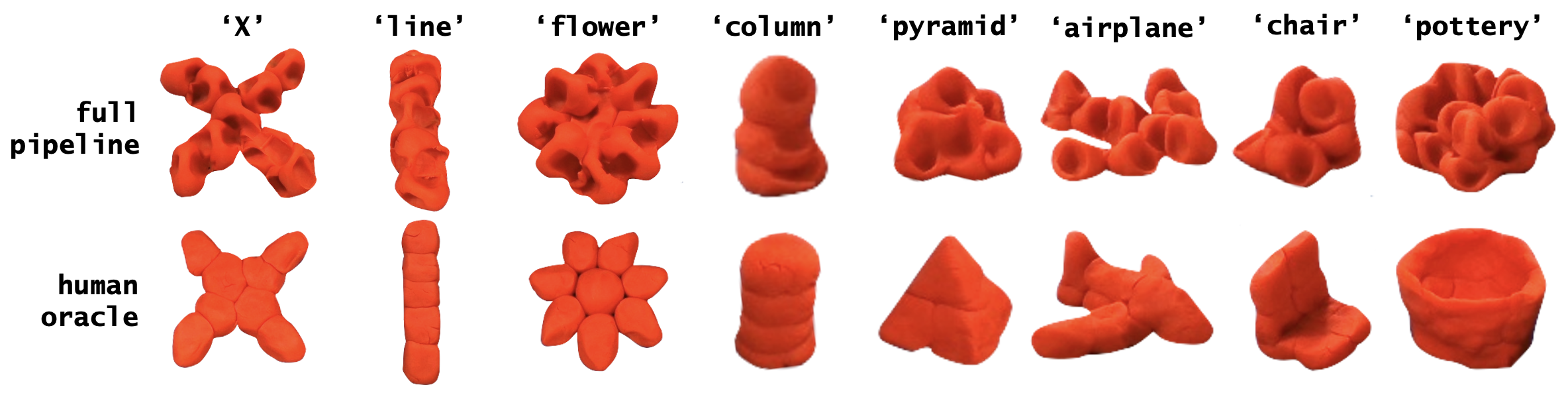}
      \caption{\label{fig:full_results} The human oracle is required to follow the same process of coarse-to-fine sculpting using their hands. The choice of camera orientation for each shape was to best visualize the full sculpture (i.e. top-down versus isometric viewpoint).} 
      \label{figurelabel}
    \end{figure*}


    \subsection{Text-to-3D Shape Goals}
    \label{sec:texto3d}


    To fully evaluate our proposed system, we conducted three real-world experiments for each shape prompt for the 'X', 'line', 'flower', 'column', 'pyramid', 'airplane', 'chair', and 'pottery' shape goals. A visualization of the final shapes created by our system is shown in Figure \ref{fig:full_results}. As we present the first text-to-3D real-world shaping system, there are no existing baselines to compare to, thus we focus on comparisons to a human oracle. The quantitative results of the final CLIP cosine similarity, an LLM shape rating prompt using the same point cloud-based representation as our sub-goal module, and human evaluations for our system compared to a human oracle are shown in Table \ref{tab:pipeline_eval}. The shapes created by our pipeline are often visually similar to that of the human oracle, but with some imperfections such as 
    indentations from the gripper. We can see from the image-based CLIP scores, the human oracle creates higher quality shapes than our proposed method (Welch's t-test, $p=7.37 e^{-6}$, $t=5.05$), though in absolute CLIP value our proposed method performs comparably for the 'X', 'line', and 'flower' shapes. Similarly, the LLM consistently rated the human oracle point clouds higher than the point clouds of our method (Welch's t-test, $p=0.77$, $t=0.29$). These patterns hold for human evaluation scores as well, with the human oracle being rated higher in both prompt similarity (Welch's t-test, $p=1.47 e^{-13}$, $t=8.14$) and shape quality (Welch's t-test, $p=1.04 e^{-11}$, $t=7.32$). However, we did find that human evaluators rated the 'X' higher than the human oracle 'X', though generally the evaluators favored the human oracle-created shapes. While there is a clear statistical significance between the human oracle and full pipeline performance on both human rating metrics, the human oracle itself did not receive a near-perfect score. The survey respondents struggled to quantify the shapes 'column' and 'chair', and to a lesser extent 'airplane' and 'pottery'.  It is very difficult to quantify the concept of how much a shape is a particular shape. To better investigate this difficulty in human evaluator consistency, we conducted a Krippendorff analysis \cite{hayes2007answering} of the inter-annotator agreement across all shapes created by both the human oracle and our pipeline. For the ordinal data of our individual Likert-type questions, the alpha scores for the shape quality and prompt similarity were $\alpha_{quality} = 0.533$, and $\alpha_{prompt} = 0.544$, respectively. For the Krippendorff analysis, a score of 0 signifies that the level of agreement is consistent with randomly answering the questions, while a score of 1 means the reviewers perfectly agreed. The scores for both of our questions were approximately 0.5, which demonstrates that the responses are not random, but the ratings have substantial noise and the reviewers are not able to fully agree. This further emphasizes the necessity to explore better strategies to quantify performance for the clay shaping task in future work, as even human evaluations have disagreement about human-created shapes.

    \subsection{Semantically Tuning Shape Goals}
    \label{sec:semantic}

\begin{figure}
      \centering
      \includegraphics[width=1.0\linewidth]{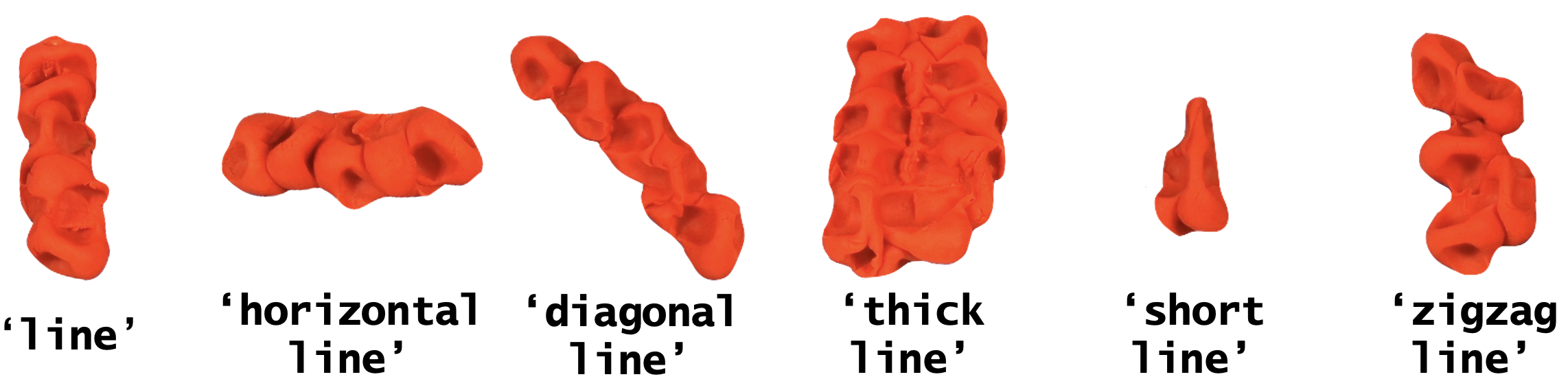}
      \caption{\label{fig:semantic} By semantically tuning the prompt, our proposed system is able to adapt the final sculpture it creates.}
      \label{figurelabel}
    \end{figure}

    Given that we present a text-to-3D sculpting system, we conducted a set of small experiments exploring how we can alter the text goal to create semantically meaningful and distinct variants of the same shape. We compare 6 variants of the 'line' shape in Figure \ref{fig:semantic}. This demonstrates the degree in which our proposed framework can adapt the final shape it is creating based on the user prompt. These more complex prompts further motivate the need for future work developing better quantitative text and 3D similarity metrics for this clay sculpting task. 

\section{Conclusion}

    In this work, we present the first fully autonomous coarse-to fine and text-to-3D clay sculpting method. While our framework is not yet able to match human performance for this complex text-to-3D task, it is able to create a range of visually identifiable shapes, and even enable semantically tuning the final sculpture output. The primary limitation of the shapes created by our proposed system is that of surface roughness and some imperfections that remain after refinement actions. Additionally, the placement strategy for the individual chunks of clay limits the variety of final shapes our system is able to create, as there must always be supporting pieces below. Through our results and human evaluation analysis, we believe that the key areas to explore for future work are to build more complex robot end-effectors to enable better finishing strategies (such as smoothing the clay surface) and to develop more reliable quantitative metrics to evaluate the quality of the final shapes created by these systems without human evaluations.


\section{Acknowledgments}

    The authors would like to thank Graham Todd for his assistance with the statistical evaluations of the results.



\bibliographystyle{IEEEtran}
\bibliography{ref}

\end{document}